\documentclass{article}
\usepackage[preprint]{colm2026_conference}
\usepackage[T1]{fontenc}
\usepackage{hyperref}
\usepackage{url}
\usepackage{booktabs}
\usepackage{array}
\usepackage{amsfonts}
\usepackage{amsmath}
\usepackage{amssymb}
\usepackage{microtype}
\usepackage{graphicx}
\usepackage{lineno}
\usepackage{float}
\usepackage{fancyhdr}

\hypersetup{
  hidelinks,
  hypertexnames=false,
  pdftitle={Selection, Recombination, or a Fresh Solve? A Candidate-Free Control for Single-Pass Test-Time Aggregation},
  pdfauthor={Guiv Farmanfarmaian},
  pdfsubject={Test-time aggregation, candidate-free controls, and language-model reasoning},
  pdfkeywords={test-time inference, language models, aggregation, reasoning, evaluation}
}
\title{Selection, Recombination, or a Fresh Solve? A Candidate-Free Control for Single-Pass Test-Time Aggregation}
\author{Guiv Farmanfarmaian\\
Department of Computer Science, ETH Zurich\\
Zurich, Switzerland}
\begin{document}
\ifcolmsubmission\linenumbers\fi
\maketitle
\lhead{Published at the COLM 2026 Workshop on Efficient Reasoning.}

\begin{abstract}
An aggregation pass gives a language model two things: candidate solutions and another chance to answer. We ask whether the candidates improve that attempt, and whether aggregation improves on selecting an existing answer. Our candidate-free control asks the same model to solve the problem again from the problem alone, using the same decoding settings and maximum output-token allowance. On AIME-2025 and HMMT-2025, Qwen3-4B-Instruct-2507 reaches 15.5\% accuracy when all four candidates are wrong, compared with 27.7\% for a fresh solve. When at least two candidates are correct, the ordering reverses: 88.7\% for aggregation versus 59.7\% for a fresh solve. With exactly one correct candidate, the benefit remains unresolved under the specified test. Yet plurality voting reaches 96.0\% in the multiple-correct regime without another model generation; its accuracy difference from aggregation is inconclusive (raw $p=0.180$). The all-wrong accuracy loss and multiple-correct gain survive a statistical correction for adding the second benchmark, developed after data collection, and persist when fresh responses are scored by their final answers. These findings concern one 4B model, two mathematics benchmarks, and a single aggregation pass over candidates truncated at their first answer. All-wrong recovery can arise from solving the problem afresh. The fresh-solve control tests whether candidates improve accuracy. Comparing with selection asks whether the additional generation improves on an existing answer.
\end{abstract}

\section{Introduction}\label{sec:intro}

After pretraining and post-training, a language model can still improve its answers by spending more computation at inference time \citep{snell2024scaling}. The question is how to use that computation: generate more candidate solutions, select among them, or reason further using what they contain. Aggregation is appealing because even a solution with a wrong final answer may contain useful intermediate steps. Trained aggregators such as SSA and GSR report gains over selection baselines \citep{qi2025ssa,wang2025gsr}. Recursive self-aggregation (RSA; \citealp{rsa2025}) repeatedly combines subsets of candidate solutions and reports gains over budget-matched parallel baselines, including on AIME-2025 and HMMT-2025 with the same Qwen3-4B model used here. These results motivate a closer look at what candidates contribute to an aggregation pass.

\begin{figure}[t]
\centering
\includegraphics[width=.97\linewidth]{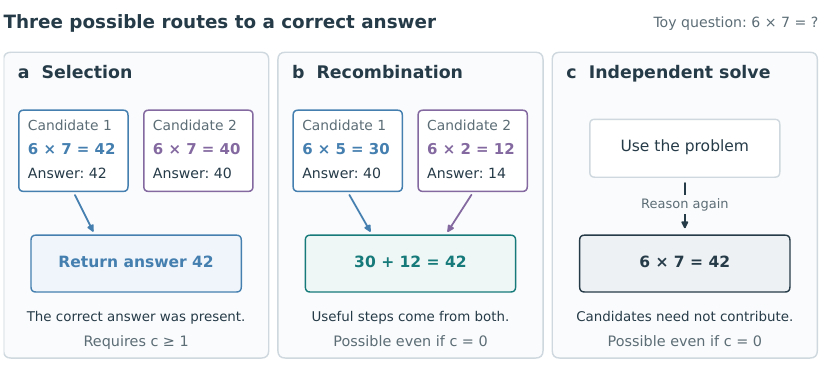}
\caption{\textbf{Three ways to reach a correct answer.} The examples are hypothetical. Selection requires a correct candidate; recombination can use steps from candidates with wrong answers; independent solving can succeed without using the displayed candidates. Here $c$ counts correct candidates. Our control measures net candidate value. Establishing the process behind an individual answer requires additional evidence.}
\label{fig:routes}
\end{figure}

Suppose a model reads four candidate solutions to a mathematics problem and then gives the correct answer. It may have selected a correct candidate, combined useful steps from several flawed ones, or solved the problem from scratch (Figure~\ref{fig:routes}). Accuracy alone cannot tell these possibilities apart.

The ambiguity is clearest when every candidate is wrong. Any method that only selects a candidate answer must fail on such a set. An aggregator's success can therefore look like evidence that it combined partial reasoning into a correct solution. The same success could also come from another attempt at solving the problem. A candidate-free comparison is needed to determine whether reading the candidates improves accuracy.

We compare two uses of the extra attempt: AGGREGATE reads the problem and candidates; NO-CANDIDATE solves the problem afresh. Both use the same model, decoding settings, and maximum output-token allowance. Their accuracy difference measures the net value of the candidate-conditioned prompt, including its instructions and length. The reasoning process remains unobserved, and total compute can differ between the arms. Our experiment tests a single prompted pass, without aggregation-specific training or recursion.

Following the candidate-correctness stratification of \citet{wang2025gsr}, we ask this question separately when none, exactly one, or at least two of four candidates are correct. Problem-averaged accuracy on all-wrong sets is 15.5\% for aggregation versus 27.7\% for a fresh solve. Recovery can therefore coexist with lower accuracy than a fresh solve. With multiple correct candidates, aggregation improves on the fresh-solve control; the one-correct regime remains unresolved under the specified test.

A second question is whether aggregation improves on choosing an existing answer. At c2+, plurality voting reaches 96.0\% versus 88.7\% for aggregation, with no additional model generation. Their accuracy comparison is inconclusive (raw $p=0.180$). At c1, aggregation is numerically well above voting. The two baselines ask whether candidates help a new attempt, and whether that attempt improves on selecting an existing answer.

\section{One extra attempt, with or without candidates}\label{sec:method}
\subsection{Design and candidate construction}

Our main model is Qwen3-4B-Instruct-2507 \citep{qwen3}. We use AIME-2025 and HMMT-Feb-2025, with 30 problems each. For each problem, we sample a pool of solutions and construct sets of $K=4$ candidates with a specified number of correct answers: none (c0), exactly one (c1), or at least two (c2+). The primary evaluation contains 472 sets, including 220 all-wrong sets from 55 problems. A problem can contribute to more than one regime. We use gold answers to construct these strata. Their sizes do not tell us how often each regime occurs in practice, and a deployed system cannot observe the regime without knowing which candidates are correct.

\begin{figure}[t]
\centering
\includegraphics[width=.97\linewidth]{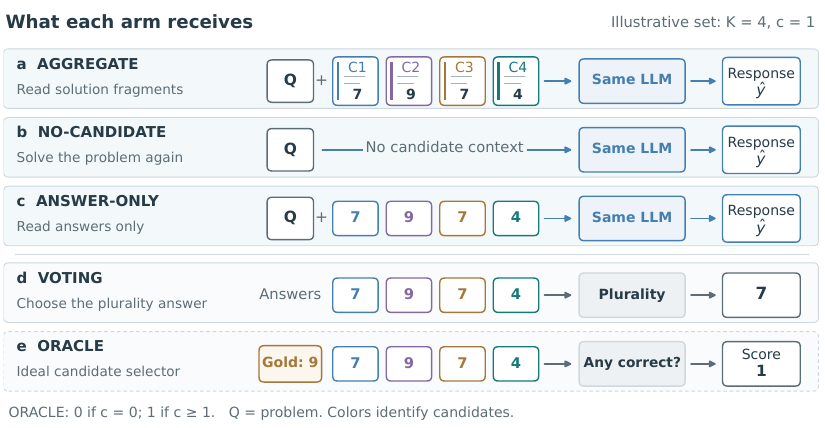}
\caption{\textbf{Five arms separate candidate information from another opportunity to generate.} AGGREGATE reads solution fragments and answers; ANSWER-ONLY reads answers; NO-CANDIDATE reads the problem alone. All three generate a response with the same model and output limit. VOTING returns an existing answer. ORACLE scores 0 at c0 and 1 at c1 or c2+. Gold answers are used only by ORACLE, scoring, and stratum construction. Every panel uses the same illustrative candidate set.}
\label{fig:arms}
\end{figure}

Each candidate shown to the model ends at its first valid extracted answer, with \texttt{<think>} content removed. A header of the form \texttt{[Candidate i] (final answer: X)} precedes the solution. Stopping at the first answer removes run-on text, but also any later self-corrections or reasoning. Our results therefore concern first-answer-truncated solution fragments. The main c0 and c2+ findings also hold when regenerated arm outputs are scored by their final answer (Section~\ref{sec:robustness}). The candidate inputs stay the same in this check; we change only how the new responses are scored. The effect of supplying complete candidate traces remains untested.

Figure~\ref{fig:arms} defines the five arms. All three generating arms use temperature 1.0, top-p 1.0, unseeded decoding, and a maximum output allowance of 16,384 tokens. NO-CANDIDATE is run for every candidate set even though it does not read the candidates. ANSWER-ONLY sees the candidate answers without their solution text. Comparing it with AGGREGATE measures what that text adds. Both arms generate a new response and can produce an answer that none of the candidates gave.

VOTING selects the plurality of raw answer strings, with a lexicographic tie-break. Equivalent answers with different surface forms can therefore split votes. ORACLE represents a perfect selector restricted to the supplied candidates: using gold correctness, it selects a correct answer whenever one is available. Its score is therefore 1 at c1 and c2+, and 0 at c0, where every available answer is wrong. This upper bound applies to selectors restricted to the supplied answers. VOTING and ORACLE both operate without another model generation after the pool exists; only VOTING is deployable.

\subsection{Estimand and statistical analysis}

Let $Y_a$ indicate answer correctness under arm $a$, let $C$ count the correct candidates in a set, and let $B$ record its regime (0, 1, or 2+). For a regime $b$, we estimate
\[
\Delta_{\mathrm{cand}}(b)=\mathbb{E}[Y_{\mathrm{AGGREGATE}}-Y_{\mathrm{NO\mbox{-}CANDIDATE}}\mid B=b],
\tag{1}
\]
where $B=2+$ means $C\ge2$. We average a problem's sets within a regime, then weight eligible problems equally. The contrast covers the full change between prompts: their instructions, length, and displayed answers, as well as the candidate reasoning.

At c0, a positive contrast would show that the candidate-conditioned prompt helps on net, without establishing recombination as the reason. A negative contrast means it lowers accuracy on average. Recombination could still help on some sets while interference hurts others. These accuracy comparisons cannot tell us how often either process occurs. We also report AGGREGATE minus ANSWER-ONLY as a secondary contrast, $\Delta_{\mathrm{reas}}$; it does not isolate a unique reasoning mechanism.

Sets from the same problem are not independent, so our primary test works at the problem level. We use an exact two-sided paired sign test on problem-level differences, dropping ties. The test asks whether problems favor one arm more often than the other. We report the mean difference in Equation~1 as a separate descriptive estimate and base our inferential claims on the corrected sign tests. Confidence intervals for the separate extraction, placebo, and structured intervention analyses appear with those results.

\textbf{Adaptive collection.} We specified the rule for adding HMMT before collecting its data. After data collection, we developed the statistical correction. The rule triggered the HMMT extension when the AIME c0 sign-test $p$-value, 0.109375, fell in the band $(0.05,0.15]$. The negative direction was not a pre-registered prediction. To account for this data-dependent collection, the correction allocates 0.025 to a possible c0 rejection on AIME alone, and 0.025 to Holm testing of the three pooled candidate contrasts. Given valid fixed-stage sign-test $p$-values, the union bound keeps the probability of any false rejection in this procedure at or below 0.05. Table~\ref{tab:main} reports the corrected decisions; Appendix~\ref{app:adaptive} gives the details and calibration.

\textbf{Answer scoring.} We re-evaluate correctness against the gold answer using the first valid answer extracted from each response. One extractor for all arms can still affect them differently: an aggregator may mention a candidate's answer before giving its own. We retain the original scoring rule for the main comparison and examine this concern by regenerating outputs and scoring each by both first and final answer (Section~\ref{sec:robustness}).

\section{Results}\label{sec:results}
We compare aggregation with a fresh solve to estimate the net accuracy effect of the candidate-conditioned prompt, and with voting to assess the benefit of generating another answer.

\subsection{When does aggregation improve on a fresh solve?}\label{sec:cand-cond}

\begin{figure}[H]
\centering
\includegraphics[width=.98\linewidth]{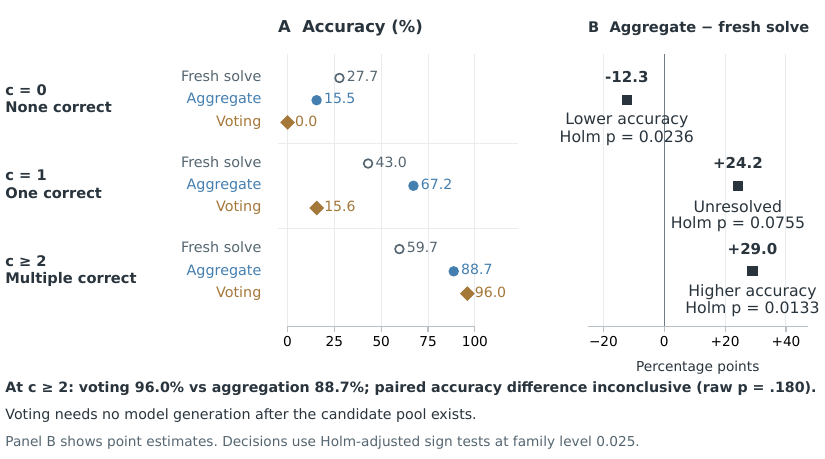}
\caption{\textbf{Aggregation, a fresh solve, and voting by candidate correctness.} Left: pooled problem-level accuracies. At c2+, voting reaches 96.0\% versus 88.7\% for aggregation and requires no generation after the pool exists; their paired accuracy comparison is inconclusive (raw sign-test $p=0.180$). Right: mean AGGREGATE minus NO-CANDIDATE, with Holm-adjusted $p$-values and decisions at family level 0.025. These sign tests assess directional consistency across problems; c1 remains unresolved despite its positive mean. Generating arms share a 16,384-token output limit. Each set contains four first-answer-truncated solution fragments.}
\label{fig:headline}
\end{figure}

Aggregation lowers accuracy by 12.3 percentage points when all candidates are wrong (c0), and raises it by 29.0 points when at least two are correct (c2+), relative to a fresh solve (Figure~\ref{fig:headline}; Table~\ref{tab:main}). With exactly one correct candidate (c1), the mean estimate is positive, but the benefit remains unresolved under the specified procedure (adjusted $p=0.0755$).

At c0, AGGREGATE reaches 15.5\% problem-averaged accuracy despite having no correct answer to select. These successes demonstrate recovery, but NO-CANDIDATE reaches 27.7\%: supplying candidates lowers accuracy overall. At c2+, the improvement over a fresh solve is compatible with selecting an available correct answer. Attributing it to recombination requires further evidence.

The c0 and c2+ effects pass the conservative adaptive correction. The c0 decision is narrow: its Holm-adjusted $p=0.0236$ is just below the stage-2 family level of 0.025. Benchmark-specific differences agree in direction: c0 is $-0.130$ on AIME and $-0.116$ on HMMT; c2+ is $+0.310$ and $+0.250$. The AIME c0 result alone is exploratory; the pooled result carries the adaptive correction. Of 55 eligible c0 problems, approximately 20 have different average accuracies between the arms and contribute to the sign test; ties are excluded. This small sample limits the empirical conclusion.

\begin{table}[H]\centering
\caption{Primary pooled results. AGG, Fresh, and Vote are problem-level accuracies. $\Delta_{\mathrm{cand}}$ and both $p$-value columns compare AGG with Fresh. The exact two-sided sign tests are Holm-adjusted across three regimes at stage-2 family level 0.025: c0 and c2+ pass; c1 remains unresolved. The voting comparison is separate (Section~\ref{sec:cand-cond}). Values are shown as fractions and rounded independently.}
\label{tab:main}\small\setlength{\tabcolsep}{7pt}
\begin{tabular}{lcccccc}\toprule
Regime & AGG & Fresh & Vote & $\Delta_{\mathrm{cand}}$ & Raw $p$ & Holm $p$\\\midrule
c0  & 0.155 & 0.277 & 0.000 & $-0.123$ & 0.0118 & 0.0236\\
c1  & 0.672 & 0.430 & 0.156 & $+0.242$ & 0.0755 & 0.0755\\
c2+ & 0.887 & 0.597 & 0.960 & $+0.290$ & 0.0044 & 0.0133\\
\bottomrule\end{tabular}
\end{table}

\textbf{Does aggregation improve on selection?} At c2+, VOTING scores 96.0\%, compared with AGGREGATE's 88.7\%, and requires no additional model generation once the candidate pool exists. Its observed advantage is 7.3 percentage points, but the paired accuracy comparison is inconclusive (10 versus 4 non-tied problems, raw $p=0.180$). This descriptive voting comparison is separate from the corrected primary tests against a fresh solve.

At c1, the ordering reverses: AGGREGATE scores 67.2\%, versus 15.6\% for VOTING. ORACLE reaches 100\% at c1 and c2+ using gold answers; it does not represent a practical selector. We have not tested a learned selector.

\subsection{Robustness checks}\label{sec:robustness}

\textbf{A simple test of recombination.} In a constructed validation task, the answer is the sum of two values, each supplied by a different wrong candidate. AGGREGATE reaches 0.767 accuracy; NO-CANDIDATE and ANSWER-ONLY both score 0.000. The 0.767 gap exceeds the pre-specified 0.30 validation margin. The experiment can detect this simple additive form of recombination when candidates supply the needed information. This check does not establish sensitivity to all forms of natural recombination or its frequency on the mathematics benchmarks (Appendix~\ref{app:gate}).

\begin{table}[t]\centering
\caption{Extraction audit on fresh generations from the same process. These are separate estimates from the original run in Table~\ref{tab:main}. Values are accuracy differences. The c0 and c2+ candidate contrasts retain their signs and reported Holm decisions. The c2+ reasoning contrast loses its reported significance under final-answer scoring. Detailed truncation and extraction checks are in Appendix~\ref{app:sensitivity}.}
\label{tab:scoring}\small
\begin{tabular}{lcc}\toprule
Contrast & First-answer scoring & Final-answer scoring\\\midrule
c0: AGGREGATE $-$ NO-CANDIDATE & $-0.118$ & $-0.159$\\
c2+: AGGREGATE $-$ NO-CANDIDATE & $+0.290$ & $+0.242$\\
c2+: AGGREGATE $-$ ANSWER-ONLY & $+0.161$ & $+0.008$\\
\bottomrule\end{tabular}
\end{table}

\textbf{Scoring first and final answers.} The original arm texts were not retained, so we generated 1,416 new responses over the same 472 sets and three generating arms. This rerun checks whether the findings recur on the same benchmark sample. Each response is scored using the original first-answer rule and an alternative that takes the last boxed answer, falling back to an answer tag, a final-answer cue, or the last line. Table~\ref{tab:scoring} reports these new estimates separately.

Under final-answer scoring, AGGREGATE minus NO-CANDIDATE is $-0.159$ at c0 and $+0.242$ at c2+; both retain their reported Holm decisions. The lower c0 accuracy also persists after excluding responses that reach the token limit and on the 178 sets whose candidates are wrong under both scoring rules ($\Delta_{\mathrm{cand}}=-0.099$, $p=0.013$; Appendix~\ref{app:sensitivity}). Candidate inputs remain truncated at their first answer. However, the c2+ advantage over ANSWER-ONLY falls from $+0.161$ to $+0.008$ under final-answer scoring. We therefore make no claim of a robust benefit from candidate reasoning beyond candidate answers in that regime.

\section{Why can wrong candidates reduce accuracy?}\label{sec:mechanism}
Aggregation is less accurate than a fresh solve when all candidates are wrong. We investigate whether displayed answers steer outputs and whether unrelated solutions can also lower accuracy.

\subsection{Does the model repeat the displayed wrong answers?}
On these all-wrong sets, among wrong outputs with an extracted answer in the original run, the fractions matching a candidate answer are 0.845 for AGGREGATE and 0.371 for NO-CANDIDATE on AIME, and 0.892 versus 0.506 on HMMT. NO-CANDIDATE never sees the candidates, so its matching rate measures coincidental overlap with the same answers. Matching alone cannot establish why accuracy falls. Across problems, matching is not significantly correlated with aggregation's accuracy relative to a fresh solve ($\rho=-0.23$).

We next replace the candidates with solution fragments from other problems, matched for length, to test whether the accuracy loss requires relevant context. This off-topic control scores 0.164 below a fresh solve (95\% problem-clustered percentile interval [$-0.241,-0.091$]). Same-problem candidates score 0.045 above this placebo ([$-0.018,+0.118$]). Of 192 wrong placebo outputs, 104 match a displayed off-topic answer. Unrelated solutions also lower accuracy. Their inconclusive comparison with same-problem candidates does not establish equivalence. Context length, task framing, displayed answers, and their interactions remain possible contributors (Appendix~\ref{app:placebo}).

\subsection{Does changing a candidate answer change the output?}\label{sec:answer-intervention}
We test the effect of displayed answers by reformatting the 220 c0 sets from 55 problems. Each candidate's terminal answer moves from its retained reasoning into a separate field. Three conditions keep the reasoning identical (Figure~\ref{fig:field-layout}). Masking tests whether hiding the answer value improves accuracy; substitution tests whether a replacement value steers the model toward that answer.

\begin{figure}[h]
\centering
\fbox{\begin{minipage}{0.92\linewidth}
\textbf{Same problem and same retained reasoning prefix in all conditions}\\[4pt]
\begin{tabular}{@{}ll@{}}
FULL & Reported final answer: \texttt{original wrong value}\\
MASKED & Reported final answer: \texttt{[ANSWER MASKED]}\\
SUBSTITUTED & Reported final answer: \texttt{controlled wrong value}
\end{tabular}\\[3pt]
\emph{Illustrative field layout. Earlier occurrences of an answer can remain in the shared reasoning prefix.}
\end{minipage}}
\caption{The structured intervention changes one explicit answer field. It differs from the original candidate rendering; equivalence between the renderings was tested and not established.}
\label{fig:field-layout}
\end{figure}

Substitution increases the probability of outputting an injected value by 0.0682 relative to FULL outputs matching the same values (95\% interval [$+0.0227,+0.1182$]; Holm-adjusted $p=0.0425$ across two confirmatory endpoints). This supports steering by the displayed value in this format. Masking produces no measurable accuracy improvement: MASKED minus FULL is $-0.0045$ ([$-0.0500,+0.0364$]; adjusted $p=1.000$). Earlier occurrences of an answer can remain in the shared reasoning after masking.

\textbf{How does this relate to the original prompt?} To assess whether the steering result extends to the original prompt, we compare formats. FULL minus original AGGREGATE is $+0.0227$ [$-0.0182,+0.0682$]. The interval extends beyond the pre-specified $\pm0.05$ equivalence margin, so equivalence was not established. The steering result remains specific to the structured format; why aggregation with the original prompt is less accurate than a fresh solve remains open (Appendix~\ref{app:task2}).

\section{Computational cost and scope}\label{sec:scope}
\textbf{Time and computation.} The shared output limit does not equalize computation. At c0 (all candidates wrong), AGGREGATE averages 18,722 input and 1,737 output tokens; NO-CANDIDATE averages 218 input and 8,310 output tokens. Input and output tokens have different computational costs. On a separate frozen sample, median c0 call times are 13.9 seconds for aggregation and 55.8 for a fresh solve. Calls run individually (batch size 1) on the same A100, without cached prompt prefixes; timings exclude candidate generation. Token means cover 220 c0 sets; timings cover 10 sets per regime, with three repeats per arm. These measurements do not establish overall efficiency (Appendix~\ref{app:t3b}).

\textbf{Generalization.} We test one 4B model on two 30-problem mathematics benchmarks, using one candidate construction and single-pass prompted aggregation. Candidates end at their first extracted answer and carry answer headers; complete traces remain untested. The c0 sign test relies on approximately 20 problems with different arm accuracies and narrowly passes the correction developed after data collection. The scoring rerun checks stability on the existing benchmark sample. A same-family reasoning-trained variant has only three c0 problems and two c1 problems, with unequal rates of reaching the token limit without an answer; generalization to stronger reasoning models remains unresolved (Appendix~\ref{app:rlvr}). We make no claim about unrelated model families, non-mathematics tasks, trained aggregators, or iterative methods.

Possible training exposure to AIME-2025 or HMMT-2025 remains unresolved and could interact with candidate context, even in comparisons on the same problems.

\textbf{Choosing a method in practice.} Candidate correctness is unknown at deployment, and our group proportions need not match real workloads. Choosing when to aggregate, vote, or solve afresh requires observable criteria and an accuracy--cost comparison on a representative workload.

\textbf{Reproducibility.} Generation logs, analysis output, and a manual audit accompany the extraction and placebo reruns but are absent from the workshop code supplement (Appendix~\ref{app:sensitivity}). Code: \url{https://github.com/guivff/candidate-free-test-time-aggregation}.

\section{Related work}\label{sec:related}
Self-consistency selects by answer agreement \citep{wang2023selfconsistency}; best-of-N uses a verifier or selector \citep{cobbe2021verifiers}. GenSelect reasons over candidates to select an answer \citep{toshniwal2025genselect,toshniwal2026genselect2}. These approaches address useful allocation and selection questions. A selector restricted to candidate answers requires at least one correct candidate to succeed.

GSR \citep{wang2025gsr} and SSA \citep{qi2025ssa} train models to refine or aggregate parallel samples. Their gains over selection baselines support learning to use candidate information. GSR's correctness-stratified evaluation motivates our regimes. We add a separate question: within a regime, does generating with candidates outperform generating without them, using the same model and output allowance? Answering this question requires a fresh-solve comparison. Our prompted, untrained 4B setting is also one in which GSR reports limited refinement ability. The capabilities of larger trained refiners require separate evaluation.

RSA repeatedly aggregates subsets of an evolving candidate pool \citep{rsa2025}. Its iterative gains and qualitative example of useful step reuse with the same Qwen model provide evidence for the promise of aggregation. RSA already compares against budget-matched parallel baselines. Our control asks what candidate context adds to a single new generation. Its net effect can be negative in our setting even if recombination occurs in individual cases or recursive aggregation improves overall performance. RSA and test-time recursive thinking \citep{zhuang2026trt} use multiple rounds, beyond our single-pass design. Appendices~\ref{app:comparison}, \ref{app:rsa}, and \ref{app:related-extra} give the extended comparisons.

\section{Conclusion}\label{sec:conclusion}
When a model answers correctly after seeing only wrong candidates, it may have combined useful steps or solved the problem on its own. A fresh-solve control measures whether the candidate-conditioned prompt improves accuracy. In our setting, aggregation lowers accuracy relative to a fresh solve when all candidates are wrong, raises it when at least two are correct, and leaves the one-correct case unresolved. The c0 and c2+ findings persist when regenerated outputs are scored by their final answer.

The benefit at c2+ is relative to a fresh solve. Plurality voting reaches 96.0\%, compared with 88.7\% for aggregation, without another model generation after the pool exists. Their accuracy comparison is inconclusive. A fresh solve measures what candidate context adds to a new attempt. Comparing with selection asks whether that attempt improves on an answer already available. How often recombination occurs, why the original all-wrong prompt lowers accuracy, and how the methods compare in end-to-end cost remain open.

\section*{Acknowledgments}
I am deeply grateful to Amir Joudaki, my adviser and mentor, for his guidance, thoughtful feedback, and continued encouragement throughout this project.

\appendix
\section*{Guide to the appendices}
The appendices explain how the comparisons were run and what the additional checks establish. Section~2 defines the five arms and the groups based on candidate correctness. Appendix~R collects the generation settings and prompt templates. The guide below locates the details needed to assess each part of the study.

\begin{center}
\small
\begin{tabular}{p{0.62\linewidth}l}
\toprule
Reader's question & Appendices\\\midrule
How were candidates, prompts, and scores constructed? & R, A, G, H\\
How were the statistical decisions made, and where are the full results? & E, L, Q\\
What do the validation and intervention checks establish? & F, I, J, M\\
What computation and response times were measured? & N, O\\
How should the results and reasoning-trained variant be interpreted? & C, P\\
How does this study relate to other aggregation methods? & B, D, K\\
\bottomrule
\end{tabular}
\end{center}

\section{Answer extraction and candidate labels}\label{app:extraction}

We determine correctness with an audited first-answer extractor, replacing the pool's stored labels. An audit against the benchmark gold answers found no false negatives. Instruction-tuned traces continue past their first answer in 99.3\% of cases, so using the last answer would change the label for $\sim$20.8\% of traces. We apply the same first-valid-answer rule across arms. A mislabeled candidate can change both its correctness regime and whether selection can succeed. The effect of any remaining labeling errors depends on their pattern: a shared extraction rule does not guarantee that errors bring the candidate contrast closer to zero. The sensitivity analysis below tests one concrete source of unequal scoring across arms.

The figures above (99.3\%, $\sim$20.8\%) describe the candidate pool and support its correctness-regime labels. AGGREGATE receives candidates truncated at their first valid answer (Section~\ref{sec:method}). The negative c0 candidate effect therefore concerns these solution fragments; its applicability to full traces remains untested. Appendix~\ref{app:sensitivity} examines whether a model's output mentions a candidate answer before its own commitment. That registered audit generates fresh responses from the same process and scores them with both first- and final-answer rules.

\section{Extended comparison to GSR, SSA, and TRT}\label{app:comparison}

GSR reports a scaling pattern consistent with our small-model result. Its Figure 2 (``A comprehensive study of refinement gap'') shows the gap growing with model size and ``weakly correlated with the base capability.'' Its prompted, untrained aggregator shows essentially no gain over majority voting (Table 1: untrained selfRef@4 average $\Delta$ of $\pm$0.0). GSR's deployed prompt explicitly instructs the model ``Don't copy candidates, use insights selectively and reason independently.'' Our separately designed prompt still produces answers that match displayed candidate answers. Since the studies differ in more than prompt wording, testing whether alternative instructions reduce matching would require a controlled prompt intervention.

SSA's error analysis (\S{}A.6 and Table 9) reports that most correct final answers already appear among the candidates. The authors attribute most apparent cases of a correct output with no correct candidate to formatting-related extraction failures. They interpret their answer-truncation probe (\S{}A.7) as evidence of some synthesis ability. TRT's loop (generate $\to$ self-rank $\to$ select-best $\to$ update strategy) combines diversity with self-verification but does not isolate the contribution of recombination. On code, it also executes self-generated tests, so its gains incorporate an execution signal in addition to model self-assessment.

\section{What the regimes imply for deployment}\label{app:heuristic}

If the correctness regime were known, the point estimates would numerically favor plurality voting at c2+ and a fresh solve at c0. At c1, aggregation's benefit over a fresh solve remains unresolved. These comparisons cannot directly guide deployment because $c$ requires correctness against gold answers. The regimes were also deliberately constructed (Section~\ref{sec:method}), so their frequencies do not describe a real workload or establish aggregation's value averaged over that workload. Within these regimes, aggregation is less accurate than a fresh solve at c0 on problems the model could otherwise solve. The placebo control (Appendix~\ref{app:placebo}) reproduces much of this accuracy loss with off-topic solutions, leaving the benefit of filtering candidates by relevance unresolved.

\section{RSA: matched budgets and a recombination example}\label{app:rsa}

RSA matches its parallel baselines' generation budgets in its headline comparison (Table 1; $K=4$, $N=16$, $T=10$ steps, 4 seeds). On AIME-2025, Pass@1 = 73.18 $\pm$ 2.20 for RSA, 68.33 for budget-matched majority voting, and 43.91 for the base model; on HMMT-2025, the respective scores are 47.55, 35.00, and 27.17. Per-round accuracy appears only in a figure (Pass@1 improves monotonically with RSA steps, their Fig.~6), so we report the trend without assigning numerical values to individual rounds. This comparison controls the total generation allowance. Our question requires an additional comparison with a candidate-free generation.

RSA's Appendix-F case study uses Qwen3-4B-Instruct-2507, our exact model, on ``the sum of the positive divisors (including 1) of 9! that have units digit 1,'' answer 103. It combines correct intermediate steps from imperfect or partial candidate traces into a solution whose structure appears in none of the individual candidates. This is a qualitative example of recombination with additional reasoning in an iterative setting. Our single-pass result shows that, when every candidate is wrong, supplying candidates can still reduce accuracy relative to a fresh solve in the setting tested here.

\section{Original analysis plan and collection rule}\label{app:prespec}

The analysis plan was fixed before collecting the confirmatory HMMT data. It accounts for repeated sets from each problem by first averaging accuracy within each arm and correctness regime. Each problem then contributes one observation to an exact two-sided sign test, with Holm correction within each $\Delta$ family (three regimes). The committed AIME$\to$HMMT rule specified three outcomes for the AIME problem-level c0 $\Delta_{\mathrm{cand}}$ sign test: report $p<0.05$ directly; extend once to new HMMT problems if the p-value falls in (0.05, 0.15], after the verifier passes its audit against HMMT gold answers; or report only at set level if $p>0.15$. The observed $p=0.109$ triggered the planned extension. Pooling gave c0 $\Delta_{\mathrm{cand}}=-0.123$, with raw pooled $p=0.0118$ and Holm-adjusted $p=0.0236$ under the later conservative stage-2 family level $\alpha=0.025$. Appendix~\ref{app:adaptive} describes that retrospective correction.

Two observations bear on the extension. (1) The directions agree across benchmarks: selective tuning to AIME alone offers no reason to expect the same direction on a disjoint benchmark. (2) A commit predating the HMMT run records the plan. We treat this as corroborating evidence because the OpenTimestamps anchor was created after the run and establishes tamper-evidence only from that point forward. The negative c0 direction was exploratory on AIME and agreed in sign on HMMT. It was not registered as a directional prediction; the inferential claim rests on the corrected pooled analysis.

\section{Construction of the validation experiment}\label{app:gate}

This validation task checks whether the same evaluation setup can detect a known opportunity to combine candidate information. Each problem's answer is $X+Y$. One candidate supplies $X$ and another supplies $Y$; both give wrong final answers. Every set is therefore c0, and recovering the answer requires combining the two candidates. The specified success criterion requires c0 AGGREGATE to exceed both NO-CANDIDATE and ANSWER-ONLY by $\geq0.30$. The observed accuracies are 0.767 for AGGREGATE and 0.000/0.000 for the comparison arms. The setup can express and detect this additive form of recombination. Its sensitivity to other forms of natural recombination remains uncalibrated, and this success gives no lower bound on their rates.

\section{Candidate-set size and incomplete generations}\label{app:ksweep}

Increasing the number of candidates does not reverse the c0 result on AIME-2025. $\Delta_{\mathrm{cand}}$ remains negative for $K\in\{2,4,8\}$, while raw AGGREGATE accuracy falls from 0.250 to 0.222 to 0.037 (Table~\ref{tab:a2}). Table~\ref{tab:a5} counts outputs that reach the token cap without an extractable answer and candidate fragments additionally shortened at the word limit. The instruction-tuned runs on both benchmarks have counts of 0 on every arm. Only the reasoning-trained variant (RLVR) has nonzero counts: 37/240 NO-CANDIDATE and 13/240 ANSWER-ONLY generations reach the cap without an answer, versus 0 for AGGREGATE. Scoring these unfinished answers as wrong lowers the comparison arms' accuracy and can favor AGGREGATE. This fixed-cap comparison is exploratory and does not bound the effect at another output allowance.

Reaching the token cap, with or without an answer, can also differ across arms. NO-CANDIDATE generates a solution from scratch, while AGGREGATE mainly reads the supplied work. In the rerun (Appendix~\ref{app:sensitivity}), instruction-tuned truncation rates are NO-CANDIDATE 41/472 (8.7\%) and ANSWER-ONLY 20/472 (4.2\%), versus AGGREGATE 3/472 (0.6\%). All arms have zero truncations without an extractable answer (\texttt{trunc\_noans} 0/472). Continuing past an answer does not affect first-answer scoring, but reaching the cap can affect final-answer scoring. The rerun therefore reports every scoring-change statistic by truncation status and bases its primary interpretation on non-truncated rows (Appendix~\ref{app:sensitivity}). The original tables were generated from committed logs with \texttt{make\_numbers.sh} and \texttt{make\_display\_items.py}. The revised figure and table builders use values reported in those analyses.

\section{Extraction-scoring sensitivity analysis}\label{app:sensitivity}

\subsection{Concern and design}
This audit asks whether the answer-scoring rule changes the comparison between aggregation and a fresh solve. The first-valid-answer extractor can score a candidate answer mentioned early in a response instead of the model's own final answer (Section~\ref{sec:method}, Appendix~\ref{app:extraction}). For example, ``Candidate 1's answer is 50, let me check\ldots{} final answer 42'' would be scored as 50. AGGREGATE and ANSWER-ONLY see candidate answers, and AGGREGATE displays each candidate's final answer on its own line. NO-CANDIDATE receives no candidate answers. The scoring rule alone could therefore make aggregation appear less accurate when all candidates are wrong (c0), more accurate when multiple candidates are correct (c2+), and more likely to be scored as matching candidate answers. We registered the design, endpoints, thresholds, and decision rule before generating the rerun data.

The original arm texts were not retained. We therefore generated and logged 1,416 fresh responses across both benchmarks, using the primary settings: identical pools, seed-0 set composition, output budget, and $T=1.0$. These new responses test the stability of the same generation process on the existing problems; the original results remain the primary reference. Candidate inputs retain their original first-answer truncation. Two rules score each new output: the unchanged first-answer extractor and the final-answer extractor, \texttt{extract\_final\_answer\_tiered}. The latter uses the last \verb|\boxed{}| answer, then an \texttt{<answer>} tag, the last ``final answer:'' cue, or the last line, and records which rule provided the answer.

For each arm, a correctness flip occurs when changing the scoring rule changes whether a response is scored correct. The primary statistic is the difference in c0 flip rates, flip(AGG) $-$ flip(NoCand). It measures whether scoring changes correctness more often for aggregation than for a fresh solve, which receives no candidate answers. We report problem-clustered 95\% bootstrap intervals ($B=10{,}000$) and stratify by truncation. When a generation reaches the token cap before committing an answer, the final-answer extractor can fall through to \texttt{last\_line}, producing disagreement caused by termination. We report truncated rows separately and exclude them from this audit's primary claims. We manually inspected the full text of every disagreement.

The project archive retains the preregistration, analysis script, machine-generated output, and manual audit. These rerun materials are not part of the workshop code supplement. The numerical results in this appendix and Appendix~\ref{app:placebo} come from that output; the primary pipeline and its locked numbers sheet remain unchanged.

\subsection{Registered decision rule and run checks}
The registered \S6 decision rule calls the difference in flip rates ``small'' when its interval's upper bound is below both 0.05 and the committed $|\Delta_{\mathrm{cand}}|\approx0.13$. It retains the conclusion that aggregation is less accurate than a fresh solve at c0 when the estimate on non-truncated rows remains negative and significant under the same Holm procedure used for the accepted-paper headline. It retains the matching result when the AGG$-$NoCand matching difference remains positive and at least 50\% of its first-answer magnitude. The same standard applies within the c2+ $\Delta_{\mathrm{cand}}$ and c0/c2+ $\Delta_{\mathrm{reas}}$ families; an unmet criterion requires restating the affected conclusion. Interpretation also requires the first-answer rerun to reproduce the committed signs and Holm outcomes. Otherwise, the rerun is inconclusive for the primary results.

All specified run checks passed: reconstruction of the 472 candidate-set identifiers against the committed logs (0 mismatches), the committed identifier-map cross-check, verification of \texttt{prompt\_sha256} on all 1,636 rows (1,416 arm responses and 220 placebo responses), an initial launch check with no empty texts and a worst within-batch duplicate-text rate of 0.008, and re-extraction checks with 0/1,636 mismatches in text hash, extracted answer, or correctness.

\subsection{Results under the two scoring rules}
The rerun met the criterion for reproducing the original results. All six per-benchmark $\Delta_{\mathrm{cand}}$ signs match the committed run, and the same effects remain significant after Holm under first-answer scoring: c0 $\Delta_{\mathrm{cand}}=-0.118$ and c2+ $\Delta_{\mathrm{cand}}=+0.290$.

\begingroup
\renewcommand{\thetable}{H.\arabic{table}}
\setcounter{table}{0}
\begin{table}[H]\centering\small
\caption{Correctness-flip rates and extracted-answer disagreement between the scoring rules at c0, over all rows. Intervals use a problem-clustered bootstrap.}
\label{tab:h1}
\begin{tabular}{lcc}\toprule
Arm & Flip rate [95\% CI] & Answer-level disagreement\\\midrule
AGGREGATE & 0.045 [0.009, 0.091] & 0.132\\
NO-CANDIDATE & 0.077 [0.041, 0.118] & 0.377\\
ANSWER-ONLY & 0.105 [0.059, 0.155] & 0.377\\
\bottomrule\end{tabular}\end{table}

Correctness changes less often for AGGREGATE than for NO-CANDIDATE, contrary to the concern that mentioning candidate answers would make aggregation more sensitive to the scoring rule (Table~\ref{tab:h1}): flip(AGG) $-$ flip(NoCand) is $-0.032$ [$-0.086,+0.027$], or $-0.053$ [$-0.121,+0.015$] on non-truncated rows. The all-row upper bound, +0.027, is below the registered 0.05 margin. AGGREGATE changes least between the scoring rules, with at least 99\% of its final-answer extractions using the boxed tier. Table~\ref{tab:h2} summarizes the pooled problem-level results and the conclusions under Holm correction within each family.

\begin{table}[H]\centering\footnotesize
\caption{Pooled results from the scoring rerun. Accuracy contrasts are problem-level differences. The p-values are raw exact sign-test values unless labelled Holm-adjusted; conclusions follow the within-family Holm procedure. Matching rates list AGGREGATE / NO-CANDIDATE.}
\label{tab:h2}
\begin{tabular}{p{0.15\linewidth}p{0.20\linewidth}p{0.20\linewidth}p{0.30\linewidth}}\toprule
Quantity & First-answer & Final-answer & Conclusion\\\midrule
c0 $\Delta_{\mathrm{cand}}$ & \shortstack[l]{$-0.118$\\$p=0.012$} & \shortstack[l]{$-0.159$\\$p<0.0001$} & Accuracy loss remains significant, including after excluding truncations.\\
c0 $\Delta_{\mathrm{reas}}$ & \shortstack[l]{$-0.041$\\$p=0.263$} & \shortstack[l]{$-0.073$\\$p=0.115$} & Negative sign retained.\\
c2+ $\Delta_{\mathrm{cand}}$ & \shortstack[l]{$+0.290$\\$p=0.0002$} & \shortstack[l]{$+0.242$\\$p=0.0003$} & Benefit remains significant.\\
c2+ $\Delta_{\mathrm{reas}}$ & \shortstack[l]{$+0.161$\\$p=0.0013$} & \shortstack[l]{$\approx0$\\Holm $p=1.000$} & Reasoning increment becomes unresolved.\\
c0 matching & 0.885 / 0.417 & 0.874 / 0.475 & Ordering retained, with at least 50\% of the difference.\\
\bottomrule\end{tabular}\end{table}
\endgroup
\setcounter{table}{2}

The matching rates in Section~4.1 come from the original run (AIME: 0.845/0.371; HMMT: 0.892/0.506). Table~\ref{tab:h2} instead pools the new responses across benchmarks. The absolute rates differ with the run and pooling, while their ordering is unchanged.

\subsection{Interpretation and remaining sensitivity}
Under final-answer scoring, aggregation falls further below a fresh solve at c0, while the descriptive matching pattern changes little. Of 161 AGGREGATE responses marked as matching a candidate answer, 15 matches disappear under the final-answer rule. Manual review identifies 9 genuine self-corrections in which a narrated candidate answer precedes the correct boxed answer; the other 6 changed extractions remain incorrect. The accuracy loss also persists on clean-c0: the 178 sets whose candidates are wrong under both scoring rules ($\Delta_{\mathrm{cand}}=-0.099$, $p=0.013$; matching 0.877). Thus aggregation remains less accurate when both scoring rules agree that every candidate in a set is wrong.

The apparent benefit of adding candidate reasoning to candidate answers at c2+ is sensitive to scoring (Table~\ref{tab:scoring}). It is significant after Holm under first-answer scoring (+0.177 in the committed run; +0.161 in the rerun) and indistinguishable from zero under final-answer scoring. Manual review finds 24 ANSWER-ONLY c2+ rows that change from wrong to correct, all on the boxed tier. In each, the model produced a correct boxed answer, but first-answer extraction selected an earlier intermediate value. Following the registered rule, we treat this c2+ benefit from reasoning text as unresolved. The central AGGREGATE-versus-NO-CANDIDATE conclusion is unchanged.

The realized-token log also supports the resource description in Section~\ref{sec:method}. At c0, AGGREGATE reads about 18.7k prompt tokens and generates 1.7k, while NO-CANDIDATE reads about 0.2k and generates 8.3k. The arms share a maximum output-token allowance, while their realized computation differs substantially.

\section{Long-context placebo control (PlaceboLong)}\label{app:placebo}

This control tests whether accuracy also falls when the displayed solutions concern unrelated problems. AGGREGATE reads about 18.7k prompt tokens at c0, while NO-CANDIDATE and ANSWER-ONLY use short contexts. Its lower accuracy relative to a fresh solve could reflect candidate content, the burden of a long noisy context, or both. The control was registered in the same commit and underwent the same checks as Appendix~\ref{app:sensitivity} (pre-reg \S9a). At c0 only, it replaces the $K$ candidates with real first-answer-truncated traces from other problems in the same pool, choosing the closest matches by solution word count. We exclude placebo answers that happen to be correct for the target problem. A separate registered random seed keeps the main experiment's candidate sets unchanged. The resulting context has similar length and contains coherent solutions to other problems.

The c0 accuracies, aggregated at the problem level, are AGGREGATE 0.173, PlaceboLong 0.127, and NO-CANDIDATE 0.291. The two registered comparisons are the off-topic-context contrast (registered label \textbf{distraction}) = PlaceboLong $-$ NoCand = $-0.164$ [$-0.241,-0.091$] (interval excludes 0); and the on-topic increment (registered label \textbf{content}) = AGG $-$ PlaceboLong = $+0.045$ [$-0.018,+0.118$] (interval includes 0). The labels describe the comparisons; they do not isolate causal mechanisms. Off-topic candidates reproduce much of the accuracy loss, and the AGGREGATE--PLACEBO difference is not statistically distinguishable from zero. This comparison does not establish equivalence. Under the registered classification rule, this result weighs against an explanation specific to relevant candidate content.

More than half of the wrong placebo outputs, 104 of 192 (a row-level rate of 0.542), matched an off-topic answer in the prompt. Answer matching therefore remains substantial even when the displayed solutions concern another problem. The accuracy loss can likewise occur without relevant candidates. The roles of context length, presenting the task as solution evaluation, displayed answers, and their interactions remain unresolved. This secondary check leaves the primary statistics in Section~\ref{sec:results} unchanged and informs the regime discussion in Appendix~\ref{app:heuristic}. The correlation between answer matching and the AGGREGATE--NO-CANDIDATE accuracy difference remains descriptive (Section~4.1).

\begin{figure}[h]
\centering
\includegraphics[width=.92\linewidth]{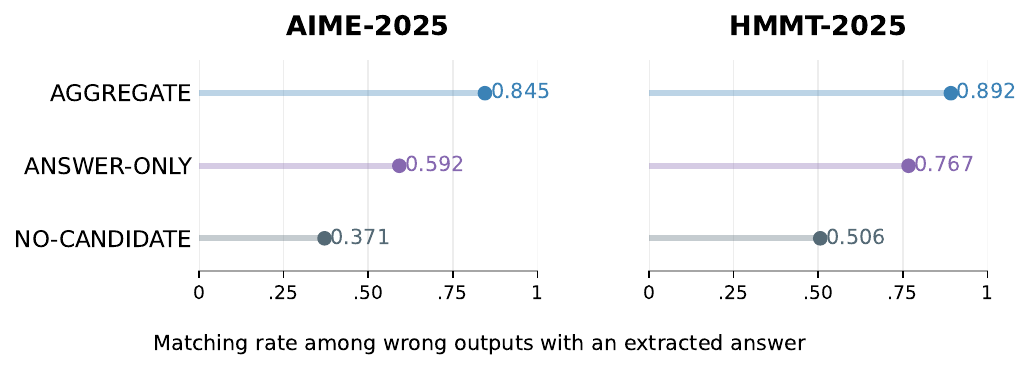}
\caption{\textbf{Descriptive matching in the original c0 experiment.} The figure shows the fraction of wrong extracted answers equal to a candidate answer displayed to AGGREGATE. These are output-level proportions for the original three generating arms; the placebo is not plotted. NO-CANDIDATE is compared with the same candidate answers although it never sees them. The original AIME paired matching test (33/4, $p<0.0001$) treats candidate sets as independent units. It is excluded from the primary inference, whose units are independent problems.}
\label{fig:matching}
\end{figure}

\section{What accuracy comparisons can establish}\label{app:identification}

The placebo arm splits the primary accuracy difference into two comparisons:
\[
\Delta_{\mathrm{cand}}(0)=(\mathrm{AGGREGATE}-\mathrm{PLACEBO})+(\mathrm{PLACEBO}-\mathrm{NO\mbox{-}CANDIDATE}).
\]
This is an algebraic identity. To interpret the first term as the benefit of relevant content and the second as the effect of context alone, we would have to assume that off-topic and genuine candidates differ only in relevance. Even under that assumption, neither term separates the individual mechanisms involved.

Average correctness also cannot reveal how often the model recombines candidate reasoning. Different mixtures of recombination, fresh reasoning, interference, and steering can produce the same arm averages. This limit concerns accuracy comparisons. The retained response texts may support future analysis of which candidate steps a response uses.

\section{Additional related work}\label{app:related-extra}

The candidate-free comparison connects to broader work on test-time scaling. Repeated sampling increases coverage predictably \citep{brown2024monkeys}, and compute-optimal test-time strategies can beat parameter scaling \citep{snell2024scaling}. Our NO-CANDIDATE arm supplies the fresh-solve comparison: another generation under the same maximum output-token allowance. Actual prompt length, output length, latency, and computation differ across arms. When matching total cost, \citet{sharma2025inverseentropy} likewise find that sequential refinement can outperform parallel self-consistency. Together, these results illustrate how the comparison and resource accounting can affect conclusions.

Self-improvement methods --- Self-Refine \citep{madaan2023selfrefine}, Reflexion \citep{shinn2023reflexion}, Tree-of-Thoughts \citep{yao2023tot}, and STaR \citep{zelikman2022star} --- ask the model to read and revise its own prior outputs. Our c0 result illustrates a related risk: in the tested setting, conditioning on wrong prior answers can reduce accuracy relative to a fresh solve. Process supervision \citep{lightman2024letsverify} and benchmark design \citep{hendrycks2021math} shape what ``a correct candidate'' means in these pools. We score final answers against gold with an audited extractor (Appendix~\ref{app:extraction}); assigning credit to individual reasoning steps remains future work.

\section{Correcting for the adaptive extension}\label{app:adaptive}

The decision to collect a second benchmark depended on the first result. The registered rule was to extend if and only if the first benchmark's problem-level c0 p-value fell in (0.05, 0.15]. The observed AIME-2025 value, $p_1=0.109375$, triggered the extension, followed by a c0 test on the pooled data. This sequence added an opportunity for rejection beyond the originally reported Holm family. We therefore developed a retrospective conservative two-stage correction after the experiment to account for the collection procedure actually used.

At stage 1, a c0 effect can be claimed from the first benchmark alone only when its raw $p_1\leq0.025$. At stage 2, the original extension rule remains unchanged, and the pooled family \{c0, c1, c2+\} is tested with Holm at family level $\alpha=0.025$. Assuming valid fixed-stage p-values, the union bound gives a family-wise error rate (FWER) $\leq0.05$ (0.025 stage-1 + 0.025 Holm stage-2). This bounds the probability of any false rejection under any dependence and any combination of true and false null hypotheses, without relying on simulation or asymptotics. We also re-derived the original values from the committed problem-level logs using the same exact sign test (stage-1 c0: 0.109375 against the reported 0.109; pooled c0: 0.011818 against the reported 0.012).

The data follow the extension path. Holm at family $\alpha=0.025$ gives the adjusted p-values in Table~\ref{tab:adaptive}: the null hypotheses for c2+ and c0 are rejected, while c1 remains unresolved. The originally reported correction at $\alpha=0.05$ gives the same two rejections and identical adjusted values. These sign tests ask whether non-tied problems favor one arm more often than the other. They do not test whether the arithmetic mean difference is zero. The c0 rejection is narrow (adjusted $p=0.0236$ against 0.025). Neither the original nor the corrected analysis claims statistical significance for c1.

We also simulated the complete procedure (stage-1 test, extension band, pooled tests, Holm) under the global null, where all tested null hypotheses hold. Each problem receives a random positive or negative sign, applied consistently across all its regimes ($B=200{,}000$; 60 problems). The original procedure's empirical family-wise error is 0.0270 (Monte Carlo 2 standard errors $\pm0.0007$) when the raw c0 p governs the stage-1 claim, and 0.0079 under a stricter reading that also requires a stage-1 Holm pass. The conservative two-stage rule gives 0.0228. This simulation calibrates error under the global null. The union-bound argument above establishes control when only some null hypotheses are true.

\begin{table}[h]\centering
\caption{The adaptive two-benchmark extension, retrospectively corrected. Stage 1 (claim from the first benchmark alone) requires raw AIME c0 $p_1\leq0.025$; the observed $p_1=0.109375$ instead entered the pre-committed extension band, so the pooled family is tested with Holm at family $\alpha=0.025$. FWER $\leq0.05$ by union bound, valid under any dependence and any partial-null configuration. The originally reported Holm correction at $\alpha=0.05$ yields the same rejections with identical adjusted values.}
\label{tab:adaptive}\small
\begin{tabular}{lcccc}\toprule
stratum & pooled $p$ & Holm-adj. $p$ & reject ($\alpha_{\mathrm{fam}}{=}0.025$) & reject ($\alpha{=}0.05$)\\\midrule
c2+ & 0.004425 & 0.0133 & yes & yes\\
c0 & 0.011818 & 0.0236 & yes & yes\\
c1 & 0.075519 & 0.0755 & no & no\\
\bottomrule\end{tabular}\end{table}

\section{The structured answer-field intervention: details}\label{app:task2}

\subsection{Design and endpoints}
This experiment asks whether a displayed candidate answer steers the model's response, and whether hiding that answer improves accuracy. We re-render all 220 eligible all-wrong sets, moving each candidate's final answer statement from its retained reasoning text to a separate ``Reported final answer'' field. The three conditions keep the retained reasoning text byte-identical:

\begin{quote}
FULL displays the original wrong answer.\\
MASKED replaces the field with \texttt{[ANSWER MASKED]}.\\
SUBSTITUTED displays a controlled wrong answer, never equivalent to gold. Substitution preserves agreement across equivalent candidate answers.
\end{quote}

The experiment yields 660 generations under the same maximum generation-token allowance with unseeded decoding. First-answer scoring is primary; the audited final-answer rule is a pre-specified sensitivity.

Endpoint A tests whether masking improves accuracy (MASKED minus FULL). Endpoint B tests whether the model follows the particular value placed in the field: the rate at which SUBSTITUTED outputs match the injected value, minus the rate at which FULL outputs match those same values. Matching in FULL provides a reference for how often those values appear without substitution. We apply Holm correction across these two confirmatory endpoints. All intervals are problem-clustered 95\% percentile confidence intervals.

\subsection{Primary results}
\begin{figure}[h]
\centering
\includegraphics[width=.82\linewidth]{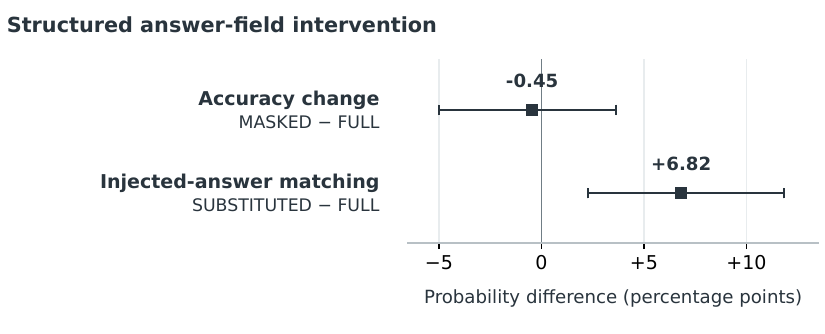}
\caption{Structured answer-field intervention on all-wrong (c0) candidate sets. All plotted values are in percentage points. Top: accuracy change from masking (MASKED $-$ FULL), $-0.45$ (95\% CI [$-5.00,+3.64$]). Bottom: value-specific steering (SUBSTITUTED output matches an injected value minus FULL output matching the same values), $+6.82$ (95\% CI [$+2.27,+11.82$]). Problem-clustered percentile bootstrap, 55 problems and 220 sets. Causal conclusions are scoped to the structured intervention.}
\label{fig:forest}
\end{figure}

\begingroup
\renewcommand{\thetable}{M.\arabic{table}}
\setcounter{table}{0}
\begin{table}[H]\centering\small
\caption{Primary intervention results under first-answer scoring. Effects are differences in probability, with 95\% confidence intervals. The raw p-values are from sign tests; Holm correction covers the two endpoints.}
\label{tab:m1}
\begin{tabular}{lccc}\toprule
Endpoint & Effect [95\% CI] & Raw $p$ & Holm $p$\\\midrule
A: masking, accuracy & $-0.0045$ [$-0.0500,+0.0364$] & 1.000 & 1.000\\
B: substitution, matching & $+0.0682$ [$+0.0227,+0.1182$] & 0.0213 & 0.0425\\
\bottomrule\end{tabular}\end{table}

The substitution effect passes the corrected test and provides evidence that changing the displayed value makes the model's extracted answer more likely to match that value within this structured format. Its sign-flip permutation test gives $p=0.0102$. The masking endpoint remains unresolved: its interval includes zero and the corrected test does not reject.

\subsection{Scoring and truncation sensitivities}
Table~\ref{tab:m2} shows how the effects change when we score final answers or exclude responses that reach the token cap. Masking the explicit field can leave earlier mentions of candidate answers in the shared reasoning. Only 2 problems have no such earlier occurrences, which is too small a subset to inform the interpretation. The retained reasoning, earlier answer mentions, distraction from long context, and interactions among these factors could all contribute to the observed effects.

\begin{table}[H]\centering\small
\caption{Intervention sensitivities. Effects are differences in probability; 95\% confidence intervals are shown where reported. These analyses use the same endpoint definitions as Table~\ref{tab:m1}.}
\label{tab:m2}
\begin{tabular}{lllc}\toprule
Endpoint & Analysis & Effect [95\% CI] & Reported $p$\\\midrule
A: masking & Final-answer scoring & +0.0182 & 0.344\\
A: masking & Truncation excluded & $-0.0091$ & 1.0\\
B: substitution & Final-answer scoring & +0.0636 [$+0.0182,+0.1136$] & 0.0352\\
B: substitution & Truncation excluded & +0.0773 [$+0.0273,+0.1303$] & 0.0213\\
\bottomrule\end{tabular}\end{table}
\endgroup
\setcounter{table}{3}

\subsection{Comparison with the original rendering}
To assess whether the structured-format finding extends to the original prompt, we compare their accuracies on the same 220 sets. FULL in the structured format exceeds original-prompt AGGREGATE by +0.0227 [$-0.0182,+0.0682$]. This interval extends beyond the pre-fixed equivalence margin of $\pm0.05$, so equivalence was not established. The causal steering result therefore remains specific to the structured format. The mechanism behind the original aggregation prompt's lower accuracy than a fresh solve remains unresolved (Section~\ref{sec:answer-intervention}).

\section{Timing with the experimental prompts}\label{app:t3b}

Table~\ref{tab:t3b_timing} reports the registered timing study using the experimental prompts, without reusing cached prompt prefixes and at batch size 1. Table~\ref{tab:task4a} reports token use from the fixed analysis artifacts. The samples differ: timing uses a frozen 30-set sample, while token means cover the full 220-set grid. We therefore report them separately. The timing estimates describe an additional generation after the candidate pool exists, under the tested execution settings. They exclude candidate generation and do not measure cached serving or end-to-end latency.

\begin{table}[H]\centering
\caption{Time per call after the candidate pool exists. The frozen 30-set sample includes 10 sets per regime, with three unseeded repeats per arm ($n=30$ per cell). Paired strategies run on the same A100 with batch size 1 and prompt-prefix caching disabled. Times use synchronized measurements; the table reports medians and empirical p95 (95th percentile) values. Token columns are medians; cap hits count outputs reaching 16,384 tokens. Candidate generation is excluded, so these measurements describe the additional call under these conditions. They do not estimate cached-serving or end-to-end cost. Table~\ref{tab:task4a} reports mean token counts on a different sample.}
\label{tab:t3b_timing}
\small
\setlength{\tabcolsep}{3.2pt}
\begin{tabular}{llrccrrr}\toprule
Group & Strategy & $n$ & \shortstack{Median\\time (s)} & \shortstack{p95\\time (s)} & \shortstack{Median input\\tokens} & \shortstack{Median output\\tokens} & \shortstack{Cap\\hits}\\\midrule
c0 & AGGREGATE-4 & 30 & 13.9 & 25.8 & 21{,}638.5 & 1{,}229.5 & 0\\
c0 & FRESH-SOLVE & 30 & 55.8 & 101.0 & 149.5 & 6{,}474.0 & 0\\
c1 & AGGREGATE-4 & 30 & 13.0 & 31.8 & 21{,}320.5 & 1{,}243.0 & 0\\
c1 & FRESH-SOLVE & 30 & 59.4 & 90.2 & 149.5 & 6{,}856.5 & 1\\
c2+ & AGGREGATE-4 & 30 & 14.3 & 23.5 & 17{,}576.0 & 1{,}437.5 & 0\\
c2+ & FRESH-SOLVE & 30 & 38.8 & 148.7 & 153.0 & 4{,}565.0 & 2\\
\bottomrule\end{tabular}
\end{table}

\begin{table}[H]\centering
\caption{Additional generation after the candidate pool exists, measured on 220 c0 sets with the model's chat template. Values are mean input and output token counts per request. They do not measure floating-point operations (FLOPs), GPU-seconds, monetary cost, or end-to-end cost. Candidate generation is also excluded. This sample differs from the 30-set timing sample in Table~\ref{tab:t3b_timing}. A complete four-candidate workflow also requires four candidate samples; their request count depends on batching.}
\label{tab:task4a}
\small
\setlength{\tabcolsep}{5pt}
\begin{tabular}{lrrr}\toprule
Strategy & \shortstack{Additional\\generation calls} & \shortstack{Mean input\\tokens} & \shortstack{Mean output\\tokens}\\\midrule
VOTE-4 (CPU plurality) & 0 & 0 & 0\\
AGGREGATE-4 & 1 & 18{,}721.7 & 1{,}737.2\\
FRESH-SOLVE (after pool) & 1 & 217.9 & 8{,}310.4\\
\bottomrule\end{tabular}
\end{table}

\section{Raw-format timing diagnostic}\label{app:timing}

We registered a batch-size-1 timing check using the original user-prompt text. After the run, we found that the timing code omitted the system message and chat template used in the main experiment. This changed when generations stopped. On the same sampled AGGREGATE sets, the experimental prompts produced a median output length of 1,366 tokens with no truncations. The raw-format requests produced a median of 16,384 tokens, and 27 of 30 reached the cap at c0. These measurements serve as an implementation diagnostic. Reported latency for the main experiment comes from the registered study using the experimental prompts without cached prefixes (Appendix~\ref{app:t3b}). The raw numerical table remains in the audit archive and supplementary provenance record.

\section{The reasoning-trained variant}\label{app:rlvr}

Among valid terminated answers, the reasoning-trained model is correct much more often (CGT 0.936 vs 0.379; Table~\ref{tab:a3}). Consequently, 200 of its 240 sets fall in c2+, leaving c0 and c1 with only three and two problems. These low-c regimes are too sparse to determine whether reasoning training changes the accuracy effect of reading candidates. The c0 effect retains its sign, but the 3-problem group is too small for reliable inference (Table~\ref{tab:a3}). In the better-populated c2+ regime, the effects retain their direction with smaller magnitudes ($\Delta_{\mathrm{cand}}+0.120$; $\Delta_{\mathrm{reas}}+0.045$). This result remains exploratory; generalization across model classes is unresolved.

The c2+ $\Delta_{\mathrm{cand}}$ also needs testing under other output allowances. NO-CANDIDATE and ANSWER-ONLY sometimes reach the cap without an answer, while AGGREGATE does not (counts: NO-CANDIDATE 37/240, ANSWER-ONLY 13/240, AGGREGATE 0; Table~\ref{tab:a5}). Such outputs score wrong, lowering those arms' accuracy and shifting AGG$-$NoCand and AGG$-$AnsOnly upward. The reported c2+ effects may therefore favor AGGREGATE more than they would under a larger allowance. The direction and size of that change remain unmeasured.

\section{Full result tables}\label{app:tables}
\begin{table}[h]\centering
\caption{Primary pooled problem-level accuracies for all five arms and the secondary contrast $\Delta_{\mathrm{reas}}=$ AGGREGATE minus ANSWER-ONLY, with its raw paired sign-test $p$-value. The primary candidate contrast and corrected inference are in Table~\ref{tab:main}. The apparent c2+ increment over ANSWER-ONLY does not persist under final-answer scoring of the process rerun (Table~\ref{tab:scoring}); it is descriptive.}
\label{tab:allarms}\small\setlength{\tabcolsep}{5pt}
\begin{tabular}{lcccccc}\toprule
Regime & AGG & Fresh & Ans-only & Vote & Oracle & $\Delta_{\mathrm{reas}}$ (raw $p$)\\\midrule
c0 & 0.155 & 0.277 & 0.186 & 0.000 & 0.000 & $-0.032$ (.359)\\
c1 & 0.672 & 0.430 & 0.539 & 0.156 & 1.000 & $+0.133$ (.041)\\
c2+ & 0.887 & 0.597 & 0.710 & 0.960 & 1.000 & $+0.177$ (.013)\\
\bottomrule\end{tabular}\end{table}
\paragraph{Reading the statistics.} The primary tests are paired sign tests at the problem level. Each problem contributes a mean accuracy per arm and c-bin. We count how many problems favor each arm and exclude ties. For example, ``10/4'' means the first arm scores higher on 10 problems and the second on 4. The reported $p$ is the two-sided exact sign-test p-value for those counts. For the primary $\Delta_{\mathrm{cand}}$ family, we apply Holm correction across the three bins at the retrospective stage-2 family level of 0.025 (Table~\ref{tab:main}). Accuracies are fractions correct under the shared, audited first-answer extractor.

\begin{table}[h]\centering
\caption{Per-benchmark accuracy by arm and $c$-bin (set-level point estimates).
Both benchmarks show the three-regime sign pattern. The
AIME$\leftrightarrow$HMMT homogeneity $p$ (permutation on per-problem
$\Delta_{\mathrm{cand}}$) is c0 0.91, c1 0.55, c2+ 0.70. Heterogeneity was
not detected, and the benchmark-specific effects agree in sign. The
pre-specified pooled estimate is reported alongside these per-benchmark
results.}
\label{tab:a1}\footnotesize\setlength{\tabcolsep}{3.5pt}
\begin{tabular}{llccccccc}\toprule
bench & $c$ & AGG & NoCand & AnsOnly & Vote & Oracle & $\Delta_{\mathrm{cand}}$ & $\Delta_{\mathrm{reas}}$\\\midrule
AIME & c0 & 0.222 & 0.352 & 0.296 & 0.000 & 0.000 & $-$0.130 & $-$0.074\\
AIME & c1 & 0.697 & 0.500 & 0.618 & 0.237 & 1.000 & $+$0.197 & $+$0.079\\
AIME & c2+ & 0.893 & 0.583 & 0.762 & 0.976 & 1.000 & $+$0.310 & $+$0.131\\
HMMT & c0 & 0.089 & 0.205 & 0.080 & 0.000 & 0.000 & $-$0.116 & $+$0.009\\
HMMT & c1 & 0.635 & 0.327 & 0.423 & 0.038 & 1.000 & $+$0.308 & $+$0.212\\
HMMT & c2+ & 0.875 & 0.625 & 0.600 & 0.925 & 1.000 & $+$0.250 & $+$0.275\\
\bottomrule\end{tabular}\end{table}

\begin{table}[h]\centering
\caption{Varying $K$ (AIME-2025 only; problem-level). Raw AGGREGATE accuracy at c0 decreases as $K$ grows, and $\Delta_{\mathrm{cand}}$ remains negative for every $K$.}
\label{tab:a2}\small\setlength{\tabcolsep}{5pt}
\begin{tabular}{lcccc}\toprule
$K$ & raw AGG c0 & $\Delta_{\mathrm{cand}}$ c0 & $\Delta_{\mathrm{cand}}$ c1 & $\Delta_{\mathrm{cand}}$ c2+\\\midrule
2 & 0.250 & $-$0.170 (.092) & $+$0.250 (.118) & $+$0.333 (.007)\\
4 & 0.222 & $-$0.130 (.109) & $+$0.197 (.302) & $+$0.310 (.022)\\
8 & 0.037 & $-$0.188 (.031) & $+$0.071 (1.00) & $+$0.274 ($<.001$)\\
\bottomrule\end{tabular}\end{table}

\begin{table}[h]\centering
\caption{RLVR (Qwen3-4B-Thinking) by $c$-bin. 200 of 240 sets fall in c2+ (25
of 30 problems); c0/c1 contain 3/2 problems, leaving insufficient power to
assess model-class differences. CGT (correct given a valid terminated
answer): instruct $244/644{=}0.379$ vs RLVR $395/422{=}0.936$.}
\label{tab:a3}\footnotesize\setlength{\tabcolsep}{3.5pt}
\begin{tabular}{lrrccccccc}\toprule
$c$ & sets & prob & AGG & NoCand & AnsOnly & Vote & Oracle & $\Delta_{\mathrm{cand}}$ (p) & $\Delta_{\mathrm{reas}}$ (p)\\\midrule
c0 & 24 & 3 & 0.083 & 0.125 & 0.208 & 0.000 & 0.000 & $-$0.042 (1.00) & $-$0.125 (1.00)\\
c1 & 16 & 2 & 0.375 & 0.438 & 0.500 & 0.000 & 1.000 & $-$0.062 (1.00) & $-$0.125 (1.00)\\
c2+ & 200 & 25 & 0.960 & 0.840 & 0.915 & 0.980 & 1.000 & $+$0.120 (.012) & $+$0.045 (.125)\\
\bottomrule\end{tabular}\end{table}

\begin{table}[h]
\centering
\caption{Validation experiment (constructed additive recombination, all c0). The pre-specified success criterion
requires AGG to exceed both comparison arms by $\geq 0.30$.}
\label{tab:a4}\small
\begin{tabular}{lc}\toprule
quantity & value\\\midrule
AGG (c0) & 0.767\\
NoCand (c0) & 0.000\\
AnsOnly (c0) & 0.000\\
validation margin: AGG $-$ max(NoCand, AnsOnly) & 0.767\\
criterion ($\geq 0.30$) & PASS\\
\bottomrule\end{tabular}\end{table}

\clearpage
\noindent\textbf{Generation completion and pool difficulty}\par\medskip

\begin{table}[H]\centering
\caption{Generation-completion checks by arm. \texttt{trunc\_noans} counts outputs that hit the token cap without an extractable answer; \texttt{cand\_trunc} counts candidate fragments additionally shortened at the word limit before aggregation. Both counts are 0 in the instruction-tuned runs. The RLVR counts occur entirely in NO-CANDIDATE (37) and ANSWER-ONLY (13 of 240). These unanswered outputs score wrong, lowering the affected arms' accuracy and shifting AGG$-$NoCand and AGG$-$AnsOnly upward. The RLVR c2+ $\Delta_{\mathrm{cand}}$/$\Delta_{\mathrm{reas}}$ therefore require testing under other output allowances before broader interpretation. The primary instruction-tuned results have no such unanswered truncations.}
\label{tab:a5}\small\setlength{\tabcolsep}{5pt}
\begin{tabular}{llrrr}\toprule run & arm & $n$ & trunc\_noans & cand\_trunc\\\midrule
instruct-AIME & AGG/NoCand/AnsOnly & 268 & 0 & 0\\
instruct-HMMT & AGG/NoCand/AnsOnly & 204 & 0 & 0\\
RLVR & AGGREGATE & 240 & 0 & 0\\
RLVR & NO-CANDIDATE & 240 & 37 & 0\\
RLVR & ANSWER-ONLY & 240 & 13 & 0\\
\bottomrule\end{tabular}\end{table}

\begin{table}[H]\centering
\caption{c0 by pool difficulty (pass@24). The c0 accuracy loss relative to a fresh solve is concentrated on problems with at least one correct pool sample. A zero pass@24 observation does not establish that a problem is unsolvable; rates on that subset remain at or near zero.}
\label{tab:a6}\small\setlength{\tabcolsep}{4.5pt}
\begin{tabular}{llrrcccc}\toprule
bench & stratum & prob & sets & AGG & NoCand & AnsOnly & $\Delta_{\mathrm{cand}}$\\\midrule
AIME & pass@24$=$0 & 9 & 36 & 0.000 & 0.000 & 0.000 & $+$0.00\\
AIME & pass@24$\geq$1 & 18 & 72 & 0.333 & 0.528 & 0.444 & $-$0.194\\
HMMT & pass@24$=$0 & 15 & 60 & 0.100 & 0.067 & 0.000 & $+$0.033\\
HMMT & pass@24$\geq$1 & 13 & 52 & 0.077 & 0.365 & 0.173 & $-$0.288\\
\bottomrule\end{tabular}\end{table}

\clearpage\section{Experimental setup and prompt reference}\label{app:setup}
\subsection{Primary generation settings}
The primary experiments use Qwen3-4B-Instruct-2507 on AIME-2025 and HMMT-2025, with 30 problems per benchmark. We first generate a pool of candidate solutions, then run the three generating arms for each set drawn from that pool. These two stages use different decoding settings. The table summarizes the study's pool records and run configuration.

\begin{center}\small
\begin{tabular}{lcc}\toprule
Setting & Candidate pool & Generating arms\\\midrule
Temperature & 0.9 & 1.0\\
Top-p & 0.95 & 1.0\\
Maximum new tokens & 8,192 & 16,384\\
Sampling seed argument & Unset & Unset\\
Generations & 24 per problem & One per set and arm\\
\bottomrule\end{tabular}
\end{center}

For the primary $K=4$ comparison, we construct four sets per eligible problem and correctness stratum with set-construction seed 0. These strata group sets by the number of correct candidate answers. Sampling is without replacement within each set; separate sets may reuse a candidate. For c2+, the construction first chooses uniformly among the feasible correct-candidate counts in $\{2,3,4\}$, then samples and shuffles the candidates. Their correctness is known when constructing these experimental strata; their frequencies do not estimate those in a natural workload (Section~2).

To prepare each candidate for display, the renderer takes the solution text after the last closing \texttt{</think>} tag, when present, and retains the fragment through its first valid answer. Candidates without a valid answer are excluded; correctness is recomputed against the gold answer. The study configures a 6,500-word limit per candidate; the original arm logs record no additional shortening from this limit. The generating arms use the vLLM backend with bfloat16 precision and a configured context limit of 49,152 tokens. The exploratory reasoning-trained model has a maximum output allowance of 32,768 tokens and is reported in Appendix~\ref{app:rlvr}.

VOTING returns the most frequent extracted answer string, breaking ties lexicographically. ORACLE scores one if any supplied candidate is correct and zero otherwise. Both operate on the existing pool and require no additional model generation.

\subsection{Prompt templates in the released implementation}
The templates below are reproduced from the released implementation (commit \texttt{3da72e7}) and agree with the corresponding study and extraction-rerun code. Placeholders for problems and candidate content are filled at runtime; the instruction \verb|\boxed{answer}| is literal. Line wrapping here is adjusted for readability. The backend combines the shared system message and each user prompt using the model tokenizer's chat template. The original headline logs do not retain the complete rendered model inputs.

\paragraph{Shared system message.}
\begin{quote}\ttfamily
You are an expert competition mathematician. Solve carefully and end\\
with the final answer on its own line as \verb|\boxed{...}|.
\end{quote}

\paragraph{NO-CANDIDATE user prompt.}
\begin{quote}\ttfamily
Problem:\\
\{problem\}\\
Solve this problem. Put ONLY the final answer on the last line as:\\
\verb|\boxed{answer}|
\end{quote}

\paragraph{AGGREGATE user prompt.}
\begin{quote}\ttfamily
Problem:\\
\{problem\}\\
Here are \{k\} candidate solutions from independent attempts. Some may be\\
wrong. Reuse correct reasoning, fix mistakes, and work out the correct\\
solution yourself.\\
\{cands\}\\
Now produce the correct, complete solution. Put ONLY the final answer on\\
the last line as: \verb|\boxed{answer}|
\end{quote}
Each candidate block takes the following form; indices start at one and blocks are separated by a blank line:
\begin{quote}\ttfamily
[Candidate \{i\}] (final answer: \{extracted\_answer\})\\
\{retained\_solution\}
\end{quote}

\paragraph{ANSWER-ONLY user prompt.}
\begin{quote}\ttfamily
Problem:\\
\{problem\}\\
Independent attempts produced these candidate final answers (no\\
reasoning shown); some may be wrong: \{answers\}\\
Solve the problem and put ONLY the final answer on the last line as:\\
\verb|\boxed{answer}|
\end{quote}
The \texttt{\{answers\}} field joins the extracted answer strings with a comma and a space, preserving candidate order.

\subsection{Available reproduction materials}
The public supplement's reproduction guide describes the included CPU analysis scripts, per-problem records, and cached statistics for the original comparisons. Candidate pools and raw generations are not included, so the supplement alone cannot repeat generation or reconstruct the scoring, placebo, structured-intervention, and timing checks from raw outputs. Appendix~\ref{app:sensitivity} describes the separately retained scoring-rerun records.


\begin{thebibliography}{19}
\bibitem[Brown et al.(2024)]{brown2024monkeys} Brown, B., Juravsky, J., Ehrlich, R., Clark, R., Le, Q.V., R\'e, C., Mirhoseini, A. Large Language Monkeys: Scaling Inference Compute with Repeated Sampling. arXiv:2407.21787, 2024.
\bibitem[Cobbe et al.(2021)]{cobbe2021verifiers} Cobbe, K., Kosaraju, V., Bavarian, M., et al. Training Verifiers to Solve Math Word Problems. arXiv:2110.14168, 2021.
\bibitem[Guo et al.(2025)]{guo2025deepseekr1} Guo, D., Yang, D., Zhang, H., et al. DeepSeek-R1 incentivizes reasoning in LLMs through reinforcement learning. Nature, 2025.
\bibitem[Hendrycks et al.(2021)]{hendrycks2021math} Hendrycks, D., Burns, C., Kadavath, S., et al. Measuring Mathematical Problem Solving with the MATH Dataset. NeurIPS Datasets and Benchmarks, 2021.
\bibitem[Lightman et al.(2024)]{lightman2024letsverify} Lightman, H., Kosaraju, V., Burda, Y., et al. Let's Verify Step by Step. ICLR, 2024.
\bibitem[Madaan et al.(2023)]{madaan2023selfrefine} Madaan, A., Tandon, N., Gupta, P., et al. Self-Refine: Iterative Refinement with Self-Feedback. NeurIPS, 2023.
\bibitem[Qi et al.(2025)]{qi2025ssa} Qi, J., Ye, X., Tang, H., Zhu, Z., Choi, E. Learning to Reason across Parallel Samples for LLM Reasoning (SSA). arXiv:2506.09014, 2025.
\bibitem[Qwen Team(2025)]{qwen3} Qwen Team. Qwen3 Technical Report. arXiv:2505.09388, 2025.
\bibitem[Shinn et al.(2023)]{shinn2023reflexion} Shinn, N., Cassano, F., Gopinath, A., Narasimhan, K., Yao, S. Reflexion: Language Agents with Verbal Reinforcement Learning. NeurIPS, 2023.
\bibitem[Snell et al.(2025)]{snell2024scaling} Snell, C., Lee, J., Xu, K., Kumar, A. Scaling LLM Test-Time Compute Optimally Can be More Effective than Scaling Parameters for Reasoning. ICLR, 2025.
\bibitem[Sharma \& Chopra(2025)]{sharma2025inverseentropy} Sharma, A., Chopra, P. The Sequential Edge: Inverse-Entropy Voting Beats Parallel Self-Consistency at Matched Compute. arXiv:2511.02309, 2025.
\bibitem[Toshniwal et al.(2025)]{toshniwal2025genselect} Toshniwal, S., Sorokin, I., Ficek, A., Moshkov, I., Gitman, I. GenSelect: A Generative Approach to Best-of-N. arXiv:2507.17797, 2025.
\bibitem[Toshniwal et al.(2026)]{toshniwal2026genselect2} Toshniwal, S., Ficek, A., Jain, S., et al. Learning Generative Selection for Best-of-N. arXiv:2602.02143, 2026.
\bibitem[Venkatraman et al.(2025)]{rsa2025} Venkatraman, S., et al. Recursive Self-Aggregation Unlocks Deep Thinking in Large Language Models (RSA). arXiv:2509.26626, 2025.
\bibitem[Wang et al.(2023)]{wang2023selfconsistency} Wang, X., Wei, J., Schuurmans, D., Le, Q., Chi, E., Narang, S., Chowdhery, A., Zhou, D. Self-Consistency Improves Chain of Thought Reasoning in Language Models. ICLR, 2023.
\bibitem[Wang et al.(2025)]{wang2025gsr} Wang, Q., Zhao, P., Huang, S., Yang, F., Wang, L., Wei, F., Lin, Q., Rajmohan, S., Zhang, D. Learning to Refine: Self-Refinement of Parallel Reasoning in LLMs (GSR). arXiv:2509.00084, 2025.
\bibitem[Yao et al.(2023)]{yao2023tot} Yao, S., Yu, D., Zhao, J., Shafran, I., Griffiths, T.L., Cao, Y., Narasimhan, K. Tree of Thoughts: Deliberate Problem Solving with Large Language Models. NeurIPS, 2023.
\bibitem[Zelikman et al.(2022)]{zelikman2022star} Zelikman, E., Wu, Y., Mu, J., Goodman, N.D. STaR: Bootstrapping Reasoning with Reasoning. NeurIPS, 2022.
\bibitem[Zhuang et al.(2026)]{zhuang2026trt} Zhuang, Y., Singh, C., Liu, L., Shen, Y., Zhang, D., Shang, J., Gao, J., Chen, W. Test-time Recursive Thinking: Self-Improvement without External Feedback (TRT). arXiv:2602.03094, 2026.
\end{thebibliography}
\end{document}